\documentclass{article} %
\usepackage{iclr2027_conference,times}

\usepackage{amsmath,amsfonts,bm}

\def\eqref#1{equation~\ref{#1}}

\def\1{\bm{1}}

\DeclareMathAlphabet{\mathsfit}{\encodingdefault}{\sfdefault}{m}{sl}
\SetMathAlphabet{\mathsfit}{bold}{\encodingdefault}{\sfdefault}{bx}{n}

\usepackage{hyperref}
\usepackage{url}
\usepackage[table]{xcolor}
\usepackage{array}
\usepackage{microtype}
\usepackage{graphicx}
\usepackage{tcolorbox}
\tcbuselibrary{breakable,skins,raster}
\usepackage{xcolor,pifont,booktabs,nicematrix}
\usepackage{subcaption}
\usepackage{booktabs}
\usepackage{amsmath}
\usepackage{amssymb}
\usepackage{mathtools}
\usepackage{amsthm}
\usepackage{times}
\usepackage{latexsym}
\usepackage[T1]{fontenc}
\usepackage[utf8]{inputenc}
\usepackage{graphicx}
\usepackage{xspace}
\usepackage{tabularx}
\usepackage{wrapfig}
\usepackage{multirow}
\usepackage{graphicx}
\usepackage{pifont}
\usepackage{float}
\usepackage{enumitem}
\usepackage{times}
\usepackage{latexsym}
\usepackage{booktabs}
\usepackage{arydshln}
\usepackage{makecell}
\usepackage{comment}
\usepackage{cleveref}
\usepackage{placeins}
\usepackage{stfloats}
\usepackage{dashrule}
\usepackage[flushmargin]{footmisc}

\title{Revisit to Segment: Working Memory Distillation for Reasoning Segmentation}

\author{%
\begin{minipage}[t]{\dimexpr\textwidth-2\tabcolsep\relax}
\vspace{0pt}
\normalfont
\resizebox{\linewidth}{!}{%
\textbf{Cilin Yan}$^{1*}$\quad
\textbf{Yilun Qiu}$^{1*}$\quad
\textbf{Wanyang Zhang}$^{1,2}$\quad
\textbf{Rui Zu}$^{1,2}$\quad
\textbf{Xiaolong Jiang}$^{1}$\quad
\textbf{Jiayin Cai}\textsuperscript{1\ding{41}}\quad
\textbf{Yao Hu}$^{1}$
}\\[3pt]
\resizebox{\linewidth}{!}{\small Xiaohongshu$^{1}$\quad PKU$^{2}$\quad {\ttfamily $^{*}$ Equal contribution\quad \ding{41} Corresponding author}}\\[2pt]
{\small\hypersetup{pdfborder={0 0 0}}\href{https://github.com/yancilin/SWiM}{\textcolor{magenta}{\texttt{https://github.com/yancilin/SWiM}}}}
\end{minipage}%
}

\newcolumntype{C}{>{\centering\arraybackslash}X}

\newcommand{\ours}{SWiM\xspace}
\definecolor{oursgray}{gray}{0.96}

\iclrfinalcopy %
\begin{document}

\maketitle
\lhead{\small Revisit to Segment: Working Memory Distillation for Reasoning Segmentation}

\begin{abstract}

Multimodal large language models (MLLMs) have approached image segmentation by reasoning about visual content and predicting target locations. Their generated responses contain reasoning traces and localization proposals that can serve as \emph{working memory} when revisiting the same image and query. Our exploration reveals that MLLMs benefit from using this self-generated working memory as context, leading to enhanced reasoning segmentation. Motivated by this finding, we seek to strengthen the backbone model's reasoning segmentation capabilities by distilling the guidance gained from revisiting prior attempts, enabling it to benefit with or without working memory at inference time. To this end, we propose Reasoning \textbf{S}egmenter with \textbf{W}ork\textbf{i}ng \textbf{M}emory~(\textbf{SWiM}), a working-memory distillation framework for reasoning segmentation. Specifically, SWiM selects rollouts based on segmentation quality to construct working memory and uses the memory-conditioned model as a teacher. The teacher provides token-level distributional supervision along student-generated trajectories, while the student receives only the original image and query. Joint optimization of on-policy self-distillation and outcome-based reinforcement learning combines working-memory guidance with direct feedback on segmentation quality. Extensive experiments on reasoning segmentation benchmarks demonstrate that SWiM achieves state-of-the-art performance, validating the effectiveness of working-memory distillation. 

\end{abstract}

\section{Introduction}

Multimodal large language models~(MLLMs) have advanced visual understanding through their ability to reason about image content and complex queries~\citep{gemini,kimik3,qwen3vl}.
These capabilities enable segmentation models to infer target regions from descriptions that require contextual understanding~\citep{sam,rasheed2024glamm,omgllava,sam2,sam3}.
Reasoning segmentation brings these capabilities together, requiring both target identification and pixel-level delineation~\citep{lisa,pixellm,zhu2025popen,mmr,star,sun2026dr,liu2026failure,yang2026don,qian2026anchorseg}.
This task calls for approaches that connect multimodal reasoning with accurate spatial localization and mask prediction.

Existing approaches typically adapt MLLMs for reasoning segmentation through supervised fine-tuning, coupling language representations with mask prediction to translate query understanding into pixel-level outputs~\citep{lisa,pixellm}.
Subsequent work incorporates structured reasoning and visual prompting to guide target localization and segmentation~\citep{cores,rsvp}.
More recently, reinforcement learning has emerged as an effective approach to strengthening visual reasoning and target localization through outcome-based rewards~\citep{samr1,segzero,visionreasoner}.
Beyond their role in producing a final prediction, the generated reasoning traces and localization proposals capture how the model interprets the query and identifies potential target regions.
This raises an open question: 
can these prior attempts guide more effective reasoning and more accurate segmentation when the model revisits the same image and query?

\suppressfloats[t]
\begin{figure}[t]
\centering
\includegraphics[width=1\linewidth]{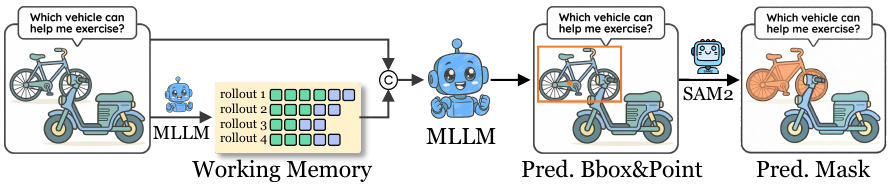}
\vspace{-3mm}
\caption{
\textbf{Working-memory-guided reasoning segmentation.}
Given an image and a query, the MLLM generates reasoning traces and localization proposals that form its working memory.
It then revisits the same image and query with this memory as additional input, predicting boxes and points that prompt a frozen SAM2 model to produce the final mask.
}
\label{fig:wm_inference}
\end{figure}

\begin{figure}[t]
\centering
\begin{subfigure}[b]{0.45\linewidth}
\centering
\includegraphics[width=\linewidth]{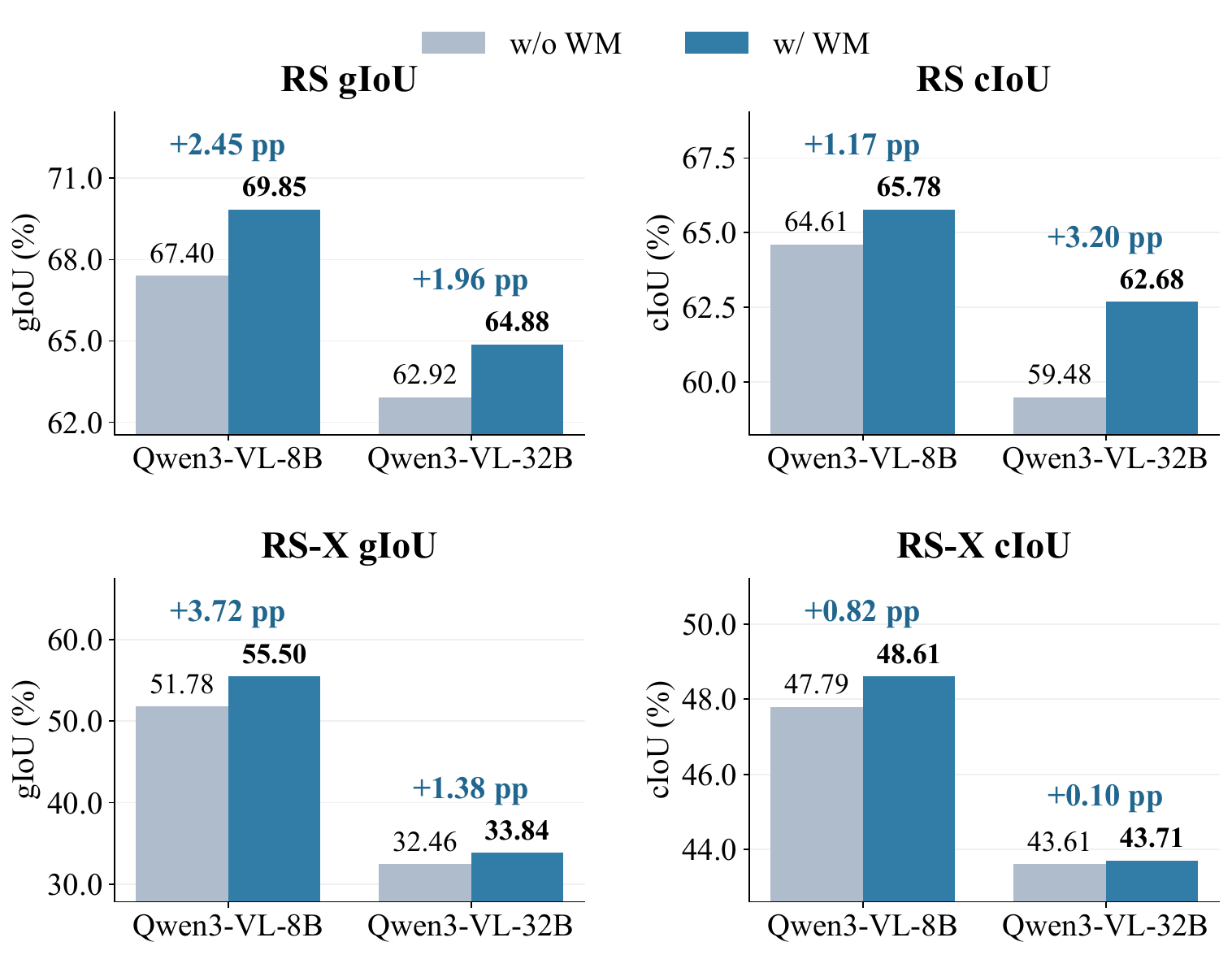}
\caption{Quantitative comparison.}
\label{fig:wm_motivation_bars}
\end{subfigure}\hfill
\begin{subfigure}[b]{0.54\linewidth}
\centering
\includegraphics[width=\linewidth]{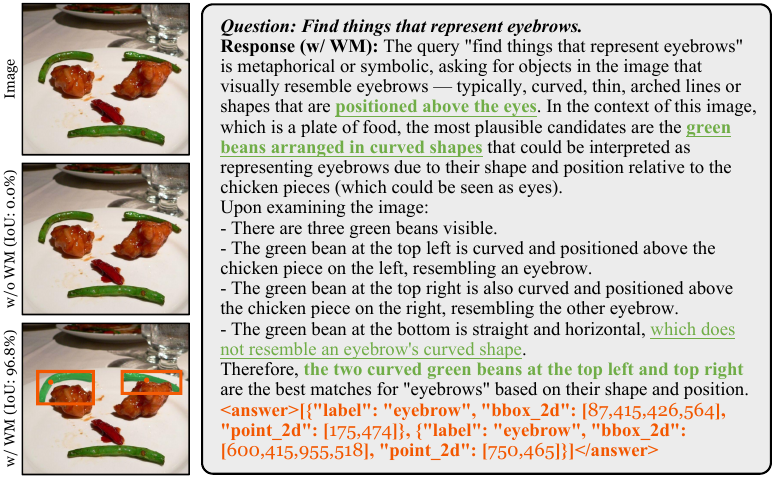}
\caption{Qualitative example.}
\label{fig:wm_motivation_case}
\end{subfigure}
\caption{\textbf{Reasoning segmentation without and with working memory (WM).} (a) Quantitative results; gains are in percentage points. (b) A qualitative example of improved target identification.}
\label{fig:wm_motivation}
\vspace{-2mm}
\end{figure}

To answer this question, we retain the model's self-generated reasoning traces and localization proposals as \emph{working memory}, which is provided as additional context when the model revisits the same image and query.
As illustrated in Figure~\ref{fig:wm_inference}, this memory enables the model to reconsider its interpretation of the query in light of its previously proposed regions.
By preserving both the reasoning and its spatial predictions, working memory provides a record of how the model approached the problem to inform a subsequent reasoning and localization attempt.
The results in Figure~\ref{fig:wm_motivation} show that conditioning on working memory consistently improves reasoning segmentation performance across model scales and benchmarks.
These improvements suggest that the information in prior attempts remains valuable beyond the predictions they initially produce.
This finding establishes self-generated working memory as a source of guidance for reasoning segmentation, allowing the model to benefit from revisiting its own reasoning and localization decisions.

Motivated by this finding, we therefore seek to convert this contextual guidance into training supervision, allowing the segmentation policy to learn from revisiting prior attempts while retaining the flexibility to predict with or without working memory at inference time.
On-policy self-distillation~(OPSD)~\citep{opsd} provides a direct way to achieve this transfer by aligning a student operating on the original image and query with a memory-conditioned teacher along student-generated trajectories.
The teacher provides token-level distributional supervision at each step of the student's own responses, translating working-memory guidance into a learning signal for reasoning and localization.
Resampling trajectories as training progresses keeps this guidance aligned with the student's evolving behavior.

To this end, we propose Reasoning \textbf{S}egmenter with \textbf{W}ork\textbf{i}ng \textbf{M}emory~(\textbf{\ours}), an on-policy self-distillation framework that transfers working-memory-guided reasoning into the segmentation policy.
At each training iteration, \ours samples rollouts from the current policy and selects responses based on segmentation quality to construct working memory.
The working-memory-conditioned model then serves as a self-teacher, providing token-level distributional supervision along student-generated trajectories, while the student receives only the original image and query.
To complement this guidance with feedback on actual segmentation outcomes, we incorporate outcome-based reinforcement learning~(RL), which directly rewards the quality of the predicted outputs.
Joint optimization of these two objectives enables the student to learn from revisiting its prior attempts while aligning its predictions with segmentation quality.
At inference time, the resulting model can retain the flexibility to predict from the image and query alone, or revisit the problem with self-generated working memory as additional context.

We conduct extensive experiments on reasoning segmentation benchmarks, where \ours achieves state-of-the-art performance.
Through further analyses, we examine the contributions of individual components and validate the effectiveness of our key design choices.

Our main contributions are summarized as follows:
\begin{itemize}[leftmargin=*, topsep=0pt, itemsep=0pt]
    \item We formulate self-generated reasoning traces and localization proposals as working memory and validate its effectiveness as both contextual information and training guidance for reasoning segmentation.
    \item To the best of our knowledge, we are the first to bring on-policy self-distillation to reasoning segmentation, where a working-memory-conditioned teacher transfers its guidance to the student through token-level supervision on student-generated trajectories.
    \item We propose \ours, a working-memory distillation framework that constructs memory from quality-selected rollouts and jointly optimizes distillation and outcome-based reinforcement learning, enabling the segmentation policy to benefit from revisiting prior attempts.
    \item \ours achieves state-of-the-art performance on reasoning segmentation benchmarks, with further analyses validating the effectiveness of our key design choices.
\end{itemize}

\section{Related Work}
\label{sec:related_work}

\textbf{Reasoning segmentation.}
Reasoning segmentation extends language-guided segmentation to queries that require contextual reasoning and world knowledge to identify the target regions.
LISA~\citep{lisa} introduces this task and connects MLLM reasoning to mask prediction by passing the projected representation of a segmentation token to a SAM mask decoder~\citep{sam}.
Early MLLM-based segmentation approaches, including PixelLM, GLaMM, and OMG-LLaVA, use supervised fine-tuning to align language representations with pixel-level outputs~\citep{pixellm,rasheed2024glamm,omgllava}.
CoReS~\citep{cores} and RSVP~\citep{rsvp} further incorporate structured reasoning and visual prompting to guide localization, emphasizing the reasoning process that precedes mask generation.
Reinforcement learning has also been adopted to strengthen visual reasoning and localization in Seg-Zero, SAM-R1, and VisionReasoner~\citep{segzero,samr1,visionreasoner}.
StAR~\citep{star} further explores reward design and rollout-based training, together with mask-level voting at inference time.

\textbf{On-policy self-distillation.}
On-policy distillation~\citep{opd} trains a student on its own generated trajectories using token-level supervision from a teacher, aligning the training distribution with the student's behavior~\citep{gkd,song2026survey,rethinkingopd}.
On-policy self-distillation extends this approach by using the model itself as a teacher conditioned on additional information~\citep{opsd}.
Existing work explores guidance from verified reasoning traces~\citep{opsd}, reusable skills~\citep{skillsd}, environment feedback~\citep{resd}, and reflections~\citep{resd}.
Multi-Rollout On-Policy Distillation~\citep{mopd} conditions the teacher on successful and failed peer rollouts from the same problem, while SEED~\citep{seed} dynamically extracts hindsight skills from on-policy trajectories for joint distillation and reinforcement learning.
\section{Problem Formulation}
\label{sec:problem_formulation}

Reasoning segmentation aims to identify and segment the target regions implicitly described by a natural-language query~\citep{lisa}.
Given an image $I$ and a query $q$, the goal is to predict a binary segmentation mask $\widehat M$ that matches the reference mask $M^\star$ for the queried regions, drawing on visual reasoning and world knowledge.

Following prior work~\citep{star}, we adopt a decoupled reasoning--segmentation formulation, where an MLLM performs reasoning and localization, and a segmentation model produces pixel-level masks.
Specifically, given an image--query pair $x=(I,q)$, the MLLM policy $\pi_\theta$ generates a response $y$ containing textual reasoning and a structured answer, from which we extract the target locations:
\begin{equation}
    y\sim\pi_\theta(\cdot\mid x), \qquad
    \mathcal{P}(y)=\{(b_j,p_j)\}_{j=1}^{n_y},
    \label{eq:reasoning_localization}
\end{equation}
where $\mathcal{P}$ extracts the localization prompts from the response, $n_y$ denotes the number of predicted targets, and $b_j$ and $p_j$ are the bounding box and representative point for the $j$-th target, respectively.
These boxes and points then serve as visual prompts for a frozen SAM2 model $S$~\citep{sam2}, which generates a mask for each target.
The final segmentation mask is their union:
\begin{equation}
    \widehat M_j = S(I,b_j,p_j), \qquad
    \widehat M = \bigcup_{j=1}^{n_y} \widehat M_j.
    \label{eq:mask_generation}
\end{equation}
\section{Methodology}
\label{sec:methodology}

\begin{figure}[t]
\centering
\includegraphics[width=1\linewidth]{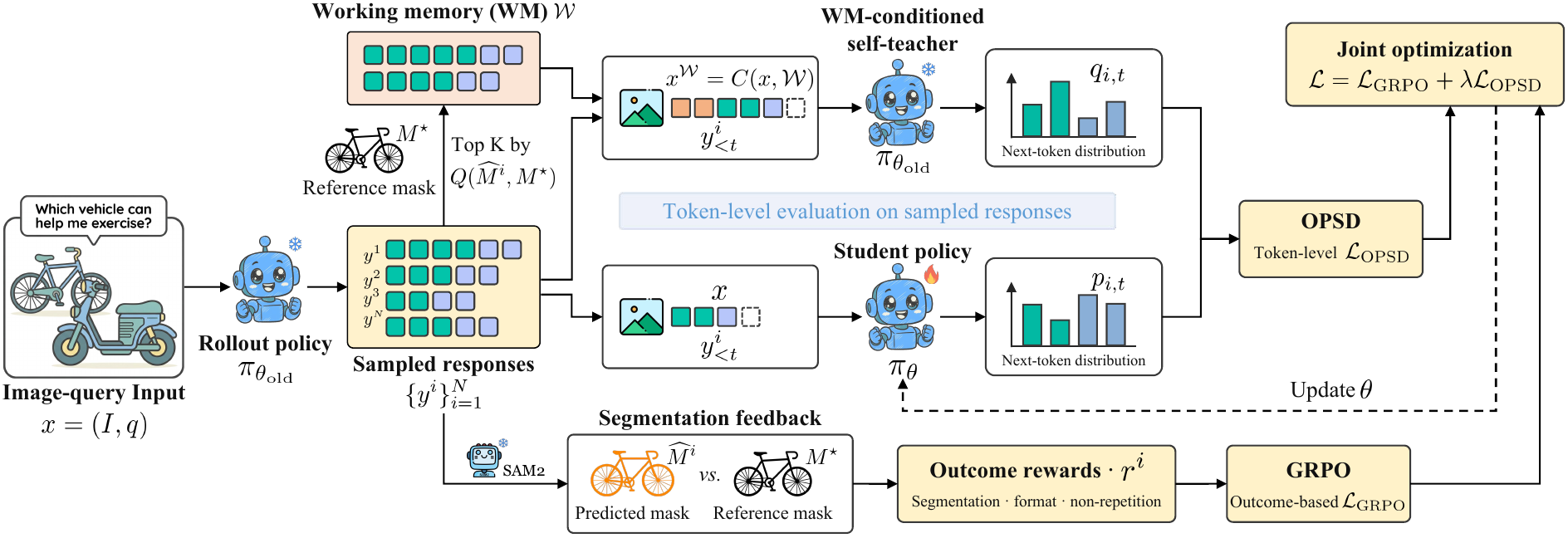}
\caption{
\textbf{Overview of \ours.}
Top-$K$ rollouts ranked by segmentation quality form working memory.
OPSD transfers token-level guidance from the working-memory-conditioned teacher to the student along the same sampled trajectories.
GRPO provides outcome-based feedback on segmentation quality.
Both objectives jointly update the student, which receives only the original image and query.
}
\label{fig:wm_method}
\end{figure}

In this section, we propose Reasoning \textbf{S}egmenter with \textbf{W}orking \textbf{M}emory~(\textbf{SWiM}), a framework for learning from revisiting self-generated working memory.
We first describe how working memory is constructed from quality-selected rollouts, followed by working memory distillation through joint optimization of on-policy self-distillation~\citep{opsd} and reinforcement learning~\citep{grpo}.
Figure~\ref{fig:wm_method} provides an overview of \ours.

\subsection{Working Memory Construction}
\label{sec:memory_construction}

We construct working memory from the policy's self-generated attempts at the same image--query pair, allowing it to revisit previous reasoning and localization proposals.
Given an input $x=(I,q)$, we sample a group of $N$ responses $\{y^{i}\}_{i=1}^{N}$ from the rollout policy $\pi_{\theta_{\mathrm{old}}}$ and obtain their segmentation masks following Section~\ref{sec:problem_formulation}.

To prioritize informative attempts, we rank the sampled responses by their segmentation quality against the reference mask $M^\star$ and retain the top $K$ as working memory $\mathcal{W}$:
\begin{equation}
    \mathcal{W}
    = \underset{y^{i}\in\{y^{1},\ldots,y^{N}\}}
      {\operatorname{TopK}}
      \; Q(\widehat M^{i},M^\star),
    \label{eq:memory_selection}
\end{equation}
where $Q$ measures segmentation quality, and $\operatorname{TopK}$ selects the $K$ top-ranked responses.

Each selected response is retained in full, preserving both its textual reasoning and structured localization outputs.
This preserves the connection between the policy's query interpretation and its proposed regions, allowing the model to revisit both when forming a new prediction.

\subsection{Working Memory Distillation}
\label{sec:memory_distillation}

Working memory makes the policy's prior reasoning traces and localization proposals available when revisiting the same input.
Building on this, we use on-policy self-distillation to transfer guidance from a working-memory-conditioned self-teacher along student-generated trajectories.
We complement this guidance with reinforcement learning driven by outcome-based rewards for segmentation predictions.
This joint optimization turns working memory into a source of token-level supervision, complemented by direct feedback on segmentation quality.

\textbf{On-policy self-distillation~(OPSD).}
At each training iteration, we construct a self-teacher from the current policy $\pi_{\theta_{\mathrm{old}}}$ by conditioning it on both the original image--query input $x=(I,q)$ and working memory $\mathcal W$:
\begin{equation}
    x^{\mathcal W}=C(x,\mathcal W),
    \label{eq:memory_context}
\end{equation}
where $C$ denotes the input constructor.
Conditioned on $x^{\mathcal W}$, the teacher can revisit prior reasoning traces and localization proposals, while the student operates solely on the original input $x$.

We reuse the responses $\{y^i\}_{i=1}^{N}$ sampled under $x$ during working memory construction.
For each response $y^i$, we evaluate the student and teacher next-token distributions at the same student-generated prefix:
\begin{equation}
    p_{i,t}=\pi_\theta(\cdot\mid x,y^i_{<t}), \qquad q_{i,t}=\pi_{\theta_{\mathrm{old}}}(\cdot\mid x^{\mathcal W},y^i_{<t}).
    \label{eq:teacher_student_distributions}
\end{equation}
Here, $y^i_{<t}$ denotes the tokens preceding position $t$, and $p_{i,t}$ and $q_{i,t}$ are distributions over the model's vocabulary $\mathcal V$.
This makes the distributions directly comparable, allowing the teacher to provide working-memory-conditioned guidance along the student's own trajectory.
The teacher distributions are computed without gradient tracking and held fixed during each policy update.

To transfer this guidance, we align the student and teacher next-token distributions using Jensen--Shannon divergence~(JSD):
\begin{equation}
    \operatorname{JSD}(p_{i,t}\,\|\,q_{i,t})=\frac{1}{2}D_{\mathrm{KL}}(p_{i,t}\,\|\,m_{i,t})+\frac{1}{2}D_{\mathrm{KL}}(q_{i,t}\,\|\,m_{i,t}),
    \label{eq:jsd}
\end{equation}
where $m_{i,t}=(p_{i,t}+q_{i,t})/2$ is the equally weighted mixture of the student and teacher distributions, and $D_{\mathrm{KL}}$ denotes Kullback--Leibler divergence.

For efficient computation, we retain the teacher's top-$L$ tokens at each position and aggregate the remaining probability mass into a single residual category.
Let $\widetilde p_{i,t}$ and $\widetilde q_{i,t}$ denote the student and teacher distributions over this shared partition.
We minimize their JSD averaged over valid response tokens:
\begin{equation}
    \mathcal L_{\text{OPSD}}=\mathbb E_{(i,t)\sim\mathcal U}\left[\operatorname{JSD}(\widetilde p_{i,t}\,\|\,\widetilde q_{i,t})\right],
    \label{eq:opsd}
\end{equation}
where $\mathcal U$ is uniform over valid token positions in the batch.
Distillation over all $N$ responses transfers working-memory guidance to the student conditioned only on the image and query.

\textbf{Reinforcement learning~(RL).}
We complement working-memory guidance with outcome-based reinforcement learning to directly optimize segmentation performance.

For each sampled response $y^i$, we obtain the segmentation prediction $\widehat M^i$ by prompting the frozen segmentor $S$ with the predicted boxes and points and taking the union of the resulting masks.
We then compute an outcome reward $r^i$:
\begin{equation}
    r^i=2r_{\text{seg}}^i+r_{\text{fmt}}^i+r_{\text{nr}}^i,
    \label{eq:outcome_reward}
\end{equation}
where $r_{\text{seg}}^i$ measures segmentation quality, $r_{\text{fmt}}^i$ evaluates the required response structure and localization fields, and $r_{\text{nr}}^i$ rewards responses
without excessive repetition.

We adopt Group Relative Policy Optimization~(GRPO) to optimize the segmentation policy.
For each group of sampled responses $\{y^i\}_{i=1}^{N}$, we compute the normalized advantage:
\begin{equation}
    A^i=\frac{r^i-\mu_r}{\sigma_r+\epsilon},
    \label{eq:grpo_advantage}
\end{equation}
where $\mu_r$ and $\sigma_r$ denote the mean and standard deviation of the group rewards, respectively, and $\epsilon$ is a small constant for numerical stability.

We define the token-level probability ratio between the student and rollout policies as
\begin{equation}
    \rho_{i,t}(\theta)=\frac{\pi_\theta(y^i_t\mid x,y^i_{<t})}{\pi_{\theta_{\mathrm{old}}}(y^i_t\mid x,y^i_{<t})}.
    \label{eq:policy_ratio}
\end{equation}
The GRPO objective is formulated as:
\begin{equation}
    \mathcal L_{\text{GRPO}}=-\mathbb E_{(i,t)\sim\mathcal U}\left[\min\left(\rho_{i,t}(\theta)A^i,\operatorname{clip}(\rho_{i,t}(\theta),1-\delta,1+\delta)A^i\right)\right],
    \label{eq:grpo}
\end{equation}
where $\delta$ is the clipping threshold.
Clipping limits the incentive for large probability changes during each update.
Optimization over all $N$ responses encourages the student to produce accurate segmentation predictions from the original image and query.

\textbf{Joint optimization.}
We jointly optimize working-memory-guided self-distillation and outcome-based reinforcement learning:
\begin{equation}
    \mathcal L_{\text{\ours}}=\mathcal L_{\text{GRPO}}+\lambda\mathcal L_{\text{OPSD}},
    \label{eq:joint_optimization}
\end{equation}
where $\lambda$ controls the strength of distillation.

Both objectives operate on the same student-generated responses: OPSD transfers token-level guidance from the working-memory-conditioned self-teacher, while GRPO provides feedback on segmentation outcomes. 
We update the student parameters $\theta$ by minimizing $\mathcal L_{\mathrm{\ours}}$, while keeping the teacher distributions and the segmentation model $S$ fixed.
At the next iteration, the updated policy generates new rollouts for working memory construction and serves as the self-teacher when conditioned on the resulting working memory.
Together, the two objectives train the policy to benefit from revisiting prior attempts while predicting from the original image and query.

\section{Experiments}
\label{sec:experiments}

\subsection{Experimental Setup}
\textbf{Benchmarks and metrics.}
We evaluate reasoning segmentation on ReasonSeg (RS)~\citep{lisa}, ReasonSeg-R (RS-R)~\citep{star}, and ReasonSeg-X (RS-X)~\citep{star}. 
These benchmarks evaluate the ability to reason about a query and segment the corresponding target regions at the pixel level.
We report gIoU, the mean of per-example intersection-over-union scores, and cIoU, the ratio of the accumulated intersection to the accumulated union.
We compute average gIoU and cIoU by taking the unweighted mean of each metric across the three benchmarks.
Additional evaluations on referring expression segmentation and multi-target segmentation are provided in Appendix~\ref{app:additional_benchmarks}.

\textbf{Baselines.}
To systematically evaluate the effectiveness of \ours, we compare it with representative reasoning segmentation methods:
(1) \emph{Supervised fine-tuning / reinforcement learning}, including LISA~\citep{lisa}, SegLLM~\citep{segllm}, READ~\citep{read}, CoReS~\citep{cores}, RSVP~\citep{rsvp}, CoPRS~\citep{coprs}, SAM-R1~\citep{samr1}, DPAD~\citep{dpad}, SAM-Veteran~\citep{samveteran}, SegCompass~\citep{segcompass}, DR$^2$Seg~\citep{dr2seg}, Seg-Zero~\citep{segzero}, VisionReasoner~\citep{visionreasoner}, LENS~\citep{zhu2026lens}, Seg-ReSearch~\citep{segresearch}, and SELF1E~\citep{self1e};
(2) \emph{Inference-time scaling / multi-round tool use}, including SAM 3 Agent~\citep{sam3}, RSAgent~\citep{rsagent}, and Rea$^2$Seg~\citep{rea2seg};
and (3) \emph{Reproduced StAR baselines}, both with and without majority voting~\citep{star}.
To ensure a fair comparison under the same training setting, we reproduce StAR without its additional training on the RS-X training split.

\textbf{Models and training.}
We instantiate \ours using backbones from the Qwen3-VL family~\citep{qwen3vl} and a frozen SAM2~\citep{sam2} model for mask generation.
For training, we begin with an RL-only warm-up that trains the policy under both plain and working-memory-augmented inputs.
Working memory is constructed online by randomly selecting up to eight preliminary responses generated by the current policy for the same image and query.
In the following working-memory distillation stage, we jointly optimize GRPO~\citep{grpo} and OPSD~\citep{opsd} on a selected training subset.
We sample $N=16$ rollouts per input, select the top $K=8$ responses by segmentation quality to construct working memory, and set the distillation weight to $\lambda=0.1$.
Detailed hyperparameters, training subset construction, and input prompts are provided in Appendices~\ref{app:implementation}--\ref{app:data_construction}.

\textbf{Evaluation settings.}
Our default evaluation setting uses a single response conditioned only on the image and query, without working memory or majority voting.
We additionally evaluate working-memory~(WM) inference, which conditions a new prediction on prior attempts, and majority voting~(MV), which aggregates multiple predicted masks.
These settings are reported separately to distinguish their inference budgets.

\subsection{Main Results}
\begin{table*}[t]
\centering
\caption{
\textbf{Performance results on reasoning segmentation benchmarks.}
``WM'' denotes working memory, and ``MV'' denotes the majority voting strategy adapted for segmentation. 
The best results in each column are highlighted in \textbf{bold}, and the second-best results are \underline{underlined}.
Higher values indicate better segmentation performance across all metrics.
}
\vspace{-2mm}
\label{tab:main_results}
\begingroup
\setlength{\tabcolsep}{3pt}
\renewcommand{\arraystretch}{1.1}
\resizebox{\textwidth}{!}{%
\begin{NiceTabular}{@{}ll*{16}{c}@{}}
\CodeBefore
  \columncolor{gray!10}{17,18}
\Body
\toprule
\multirow{3}{*}{Method} & \multirow{3}{*}{\shortstack{Base model\\(Size)}}
& \multicolumn{2}{c}{\multirow{2}{*}{RS test}} & \multicolumn{2}{c}{\multirow{2}{*}{RS-R}} & \multicolumn{10}{c}{RS-X test} & \multicolumn{2}{c}{} \\
\cmidrule(lr){7-16}
& & \multicolumn{2}{c}{} & \multicolumn{2}{c}{} & \multicolumn{2}{c}{overall} & \multicolumn{2}{c}{P/F} & \multicolumn{2}{c}{C/KI} & \multicolumn{2}{c}{C/R} & \multicolumn{2}{c}{C/MH} & \multicolumn{2}{c}{\multirow{-2}{*}{Average}} \\
\cmidrule(lr){3-4}\cmidrule(lr){5-6}\cmidrule(lr){7-8}\cmidrule(lr){9-10}\cmidrule(lr){11-12}\cmidrule(lr){13-14}\cmidrule(lr){15-16}\cmidrule(lr){17-18}
& & gIoU & cIoU & gIoU & cIoU & gIoU & cIoU & gIoU & cIoU & gIoU & cIoU & gIoU & cIoU & gIoU & cIoU & gIoU & cIoU \\
\midrule
\multicolumn{18}{l}{\emph{Supervised Fine-Tuning / Reinforcement Learning}} \\
LISA & Llama2-13B & 51.5 & 51.3 & 52.5 & 53.3 & 25.1 & 26.0 & 27.9 & 30.6 & 28.2 & 29.2 & 26.8 & 25.8 & 13.8 & 16.8 & 43.0 & 43.5 \\
SegLLM & LLaVA1.5-7B & 52.4 & 48.4 & -- & -- & -- & -- & -- & -- & -- & -- & -- & -- & -- & -- & -- & -- \\
READ & LLaVA1.5-13B & 62.2 & 62.8 & -- & -- & -- & -- & -- & -- & -- & -- & -- & -- & -- & -- & -- & -- \\
CoReS & LLaVA1.5-13B & 65.5 & -- & -- & -- & -- & -- & -- & -- & -- & -- & -- & -- & -- & -- & -- & -- \\
RSVP & GPT-4o & 60.3 & 60.0 & -- & -- & -- & -- & -- & -- & -- & -- & -- & -- & -- & -- & -- & -- \\
CoPRS & Qwen2.5-VL-7B & 59.8 & 55.1 & -- & -- & -- & -- & -- & -- & -- & -- & -- & -- & -- & -- & -- & -- \\
SAM-R1 & Qwen2.5-VL-7B & 60.2 & 54.3 & -- & -- & -- & -- & -- & -- & -- & -- & -- & -- & -- & -- & -- & -- \\
DPAD & Qwen2.5-VL-7B & 60.8 & 57.5 & -- & -- & -- & -- & -- & -- & -- & -- & -- & -- & -- & -- & -- & -- \\
SAM-Veteran & Qwen2.5-VL-7B & 62.6 & 56.1 & -- & -- & -- & -- & -- & -- & -- & -- & -- & -- & -- & -- & -- & -- \\
SegCompass & Qwen2.5-VL-7B & 64.0 & 64.8 & -- & -- & -- & -- & -- & -- & -- & -- & -- & -- & -- & -- & -- & -- \\
DR$^2$Seg & Qwen2.5-VL-7B & 66.1 & 63.6 & 67.4 & 61.4 & 45.2 & 36.6 & 52.1 & 40.9 & 55.9 & 47.3 & 43.9 & 34.1 & 23.7 & 25.7 & 59.6 & 53.9 \\
Seg-Zero & Qwen2.5-VL-7B & 57.5 & 52.0 & -- & -- & -- & -- & -- & -- & -- & -- & -- & -- & -- & -- & -- & -- \\
VisionReasoner & Qwen2.5-VL-7B & 63.6 & 55.7 & 64.8 & 56.8 & 42.2 & 33.8 & 50.1 & 45.1 & 50.7 & 39.0 & 39.4 & 31.1 & 24.5 & 23.8 & 56.9 & 48.8 \\
LENS & Qwen2.5-VL-3B & 57.2 & 58.0 & -- & -- & -- & -- & -- & -- & -- & -- & -- & -- & -- & -- & -- & -- \\
Seg-ReSearch & Qwen3-VL-8B & 67.4 & 59.0 & -- & -- & -- & -- & -- & -- & -- & -- & -- & -- & -- & -- & -- & -- \\
SELF1E & InternVL3-8B & 65.7 & \underline{67.0} & -- & -- & -- & -- & -- & -- & -- & -- & -- & -- & -- & -- & -- & -- \\
\midrule
\multicolumn{18}{l}{\emph{Inference-Time Scaling / Multi-Round Tool Use}} \\
SAM 3 Agent & Qwen2.5-VL-7B & 62.6 & 56.2 & 63.1 & 58.0 & 34.4 & 29.5 & 41.7 & 37.9 & 37.7 & 31.8 & 35.3 & 26.0 & 17.9 & 21.9 & 53.4 & 47.9 \\
SAM 3 Agent & Qwen3-VL-8B & 70.2 & \textbf{67.3} & 69.3 & 64.1 & 42.3 & 39.7 & 50.6 & 43.6 & 48.4 & 45.1 & 40.5 & 40.8 & 25.8 & 26.0 & 60.6 & 57.0 \\
SAM 3 Agent & Qwen2.5-VL-72B & \underline{71.8} & 65.2 & 72.4 & 65.3 & 49.8 & 40.3 & 57.8 & 50.5 & 55.5 & 45.4 & 49.6 & 34.0 & 30.8 & 35.1 & 64.7 & 56.9 \\
RSAgent & Qwen2.5-VL-7B & 66.5 & 57.9 & -- & -- & -- & -- & -- & -- & -- & -- & -- & -- & -- & -- & -- & -- \\
Rea$^2$Seg & Qwen2.5-VL-3B & 66.6 & 65.5 & -- & -- & -- & -- & -- & -- & -- & -- & -- & -- & -- & -- & -- & -- \\
\midrule
\multicolumn{18}{l}{\emph{Reproduced StAR Baselines (w/o RS-X Training Data)}} \\
StAR & Qwen3-VL-8B & 68.6 & 60.2 & 71.3 & 65.7 & 54.0 & 49.2 & 60.9 & 51.6 & 62.5 & 55.1 & 53.5 & 49.1 & 33.5 & 39.4 & 64.6 & 58.4 \\
StAR + MV & Qwen3-VL-8B & 69.0 & 60.8 & 72.0 & 66.7 & 55.5 & 51.1 & 63.7 & 54.5 & 61.8 & 54.9 & 54.7 & 51.4 & 37.0 & 42.1 & 65.5 & 59.5 \\
StAR & Qwen3-VL-32B & 70.8 & 66.8 & 71.8 & 67.5 & 58.6 & 54.3 & 67.0 & 61.3 & 66.7 & 64.3 & 56.3 & 49.4 & 39.9 & 42.2 & 67.1 & 62.9 \\
StAR + MV & Qwen3-VL-32B & 71.7 & 66.7 & 73.2 & 67.3 & 61.4 & 57.7 & \underline{71.6} & \underline{68.1} & \underline{68.5} & 66.9 & 58.0 & 50.8 & 43.4 & 46.4 & 68.8 & 63.9 \\
\midrule
\multicolumn{18}{l}{\emph{Our Method (w/o RS-X Training Data)}} \\
\textbf{\ours} & Qwen3-VL-8B & 70.8 & 65.8 & 73.0 & \textbf{70.8} & 55.6 & 49.3 & 62.5 & 54.9 & 65.6 & 60.4 & 54.1 & 44.4 & 35.6 & 39.8 & 66.5 & 62.0 \\
\textbf{\ours + WM\&MV} & Qwen3-VL-8B & 71.6 & \textbf{67.3} & 73.5 & \underline{70.3} & 58.1 & 51.5 & 65.4 & 57.1 & 67.0 & 63.9 & 54.8 & 45.7 & 41.9 & 42.4 & 67.8 & 63.0 \\
\textbf{\ours} & Qwen3-VL-32B & 71.5 & 66.7 & \underline{73.6} & 69.9 & \underline{62.6} & \underline{58.5} & 70.7 & 62.3 & 67.1 & \underline{67.7} & \underline{58.5} & \underline{53.8} & \textbf{52.4} & \underline{51.3} & \underline{69.2} & \underline{65.0} \\
\textbf{\ours + WM\&MV} & Qwen3-VL-32B & \textbf{72.4} & 66.2 & \textbf{74.7} & 69.5 & \textbf{64.1} & \textbf{61.9} & \textbf{73.7} & \textbf{70.6} & \textbf{69.7} & \textbf{69.4} & \textbf{60.8} & \textbf{56.1} & \underline{48.8} & \textbf{53.0} & \textbf{70.4} & \textbf{65.9} \\
\bottomrule
\end{NiceTabular}%
}
\endgroup
\vspace{-3mm}
\end{table*}

Table~\ref{tab:main_results} presents the experimental results on reasoning segmentation benchmarks, from which we draw the following observations:
\begin{itemize}[leftmargin=*, topsep=0pt, itemsep=0pt]
    \item 
    Scaling the backbone from Qwen3-VL-8B to Qwen3-VL-32B improves most metrics, particularly on RS-X, demonstrating the scalability of \ours.
    At both scales, incorporating working memory and majority voting yields further gains in average performance, highlighting the compatibility of working-memory distillation with working-memory-guided reasoning at inference time.
    \item 
    \ours achieves state-of-the-art overall performance, attaining the highest average gIoU and cIoU among the compared methods.
    When predicting from the image and query alone, without working memory or majority voting, \ours outperforms the strongest reported results in both the \emph{Supervised Fine-Tuning / Reinforcement Learning} and \emph{Inference-Time Scaling / Multi-Round Tool Use} categories on both average metrics.
    Notably, with both models predicting directly from the image and query using the same Qwen3-VL-8B backbone, \ours surpasses the reproduced StAR baseline by 1.9 and 3.6 percentage points in average gIoU and cIoU, respectively.
    These results demonstrate the effectiveness of our framework in strengthening reasoning segmentation.
\end{itemize}

\subsection{Ablation Studies}

In this section, we conduct ablation studies to assess the contributions of individual components and examine the key design choices in \ours.

\begin{figure*}[t]
\centering
\includegraphics[width=\textwidth]{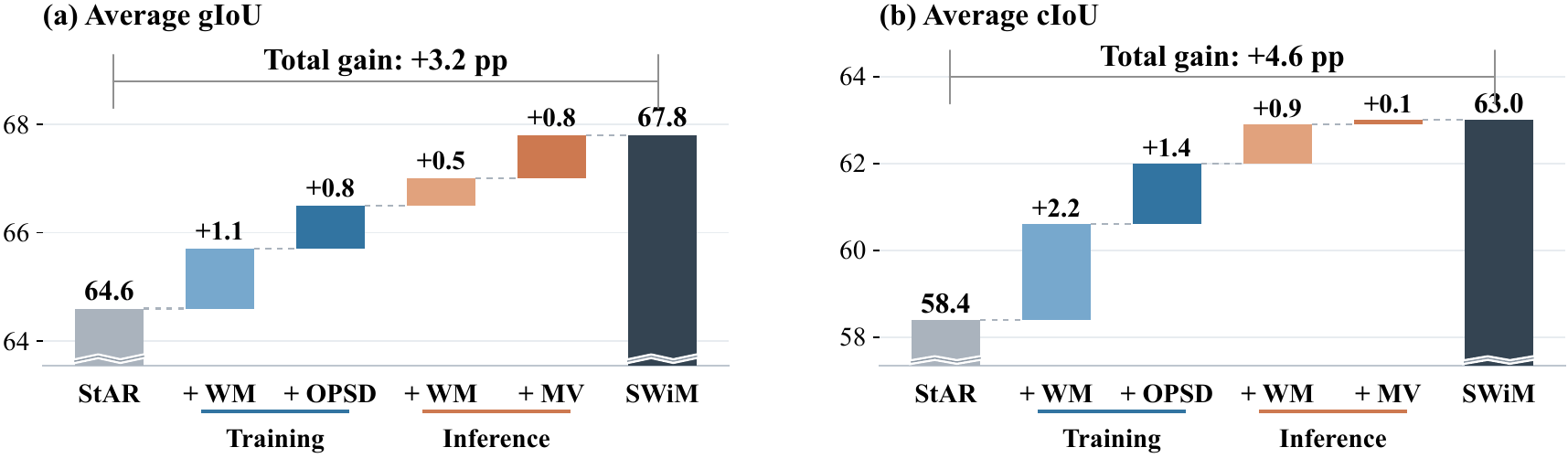}
\vspace{-6mm}
\caption{
\textbf{Effects of training and inference configurations in \ours.}
Results are averaged over RS, RS-R, and RS-X.
Floating bars show gains in percentage points; solid bars show absolute scores.
}
\vspace{-2mm}
\label{fig:swim_progression}
\end{figure*}

\textbf{Training and inference configurations.}
We investigate the contributions of working-memory-guided training, distillation, and inference through a progressive comparison of these configurations with Qwen3-VL-8B as the backbone.
As shown in Figure~\ref{fig:swim_progression}, introducing working memory during the RL warm-up improves performance.
Subsequent working-memory distillation yields further gains under plain inference, suggesting that memory-conditioned guidance can strengthen predictions from the image and query alone.
Providing working memory at inference further improves performance, and combining it with majority voting achieves the best average results.
The full configuration improves average gIoU and cIoU over the reproduced StAR baseline by 3.2 and 4.6 percentage points, respectively, supporting the benefits of working memory during both training and inference.
Detailed per-benchmark results are provided in Appendix~\ref{app:progression}.

\textbf{GRPO and OPSD in working memory distillation.}
\begin{table*}[t]
\centering

\caption{
\textbf{Ablation of training objectives.}
We compare GRPO, OPSD, and their joint optimization on RS, RS-R, and RS-X without working memory at inference.
}
\vspace{-2mm}
\label{tab:grpo_opd_ablation}
\begingroup
\renewcommand{\arraystretch}{0.8}
\begin{tabular}{cccccccccc}
\toprule
\multirow{2}{*}{GRPO} & \multirow{2}{*}{OPSD} & \multicolumn{2}{c}{RS} & \multicolumn{2}{c}{RS-R} & \multicolumn{2}{c}{RS-X} & \multicolumn{2}{c}{Average} \\
\cmidrule(lr){3-4}\cmidrule(lr){5-6}\cmidrule(lr){7-8}\cmidrule(lr){9-10}
 &  & gIoU & cIoU & gIoU & cIoU & gIoU & cIoU & gIoU & cIoU \\
\midrule

& & 69.6 & 64.4 & 72.4 & 67.6 & 55.1 & \textbf{50.0} & 65.7 & 60.6 \\
$\checkmark$ &  & 70.6 & 64.7 & 73.0 & 70.2 & 55.1 & 47.8 & 66.2 & 60.9 \\
 & $\checkmark$ & 70.1 & \textbf{65.9} & 72.3 & 68.4 & 54.8 & 48.4 & 65.7 & 60.9 \\
\rowcolor{gray!10}
$\checkmark$ & $\checkmark$ & \textbf{70.8} & 65.8 & \textbf{73.0} & \textbf{70.8} & \textbf{55.6} & 49.3 & \textbf{66.5} & \textbf{62.0} \\
\bottomrule
\end{tabular}
\endgroup
\vspace{-3mm}
\end{table*}
We examine the contributions of GRPO and OPSD by optimizing each objective independently and jointly, with all models evaluated without working memory at inference.
As shown in Table~\ref{tab:grpo_opd_ablation}, applying GRPO or OPSD alone yields modest improvements over the model after RL warm-up, while joint optimization achieves the best average performance, outperforming training with either objective alone.
These results suggest that token-level guidance from the memory-conditioned teacher complements outcome-based segmentation feedback, enabling more effective working-memory distillation and strengthening predictions from the image and query alone.

\begin{wrapfigure}{R}{0.42\textwidth}
\centering
\vspace{-5mm}
\includegraphics[width=\linewidth]{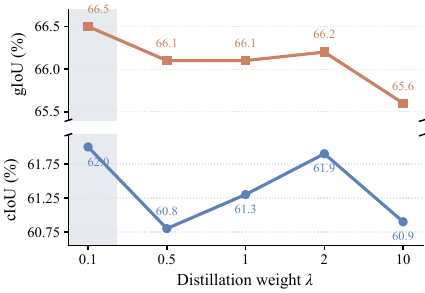}
\vspace{-6mm}
\caption{
\textbf{Effect of the distillation weight $\lambda$.}
gIoU and cIoU are averaged over RS, RS-R, and RS-X.
}
\vspace{-3mm}
\label{fig:parameter_sensitivity}
\end{wrapfigure}

\textbf{Distillation weight.}
We examine the effect of $\lambda$, which balances OPSD and GRPO in the joint training objective.
As shown in Figure~\ref{fig:parameter_sensitivity}, $\lambda=0.1$ achieves the best average gIoU and cIoU among the evaluated settings, while a high weight of $\lambda=10$ reduces performance.
These results suggest that the strength of working-memory guidance should be balanced with outcome-based optimization.
We therefore use $\lambda=0.1$ as the default setting.
Additional analysis of working memory size is provided in Appendix~\ref{app:memory_size}.

We further examine the design choices of \ours through additional experiments reported in the appendix.
The working-memory-size study in Appendix~\ref{app:memory_size} shows that larger memory does not consistently improve performance.
The teacher-context comparison in Appendix~\ref{app:teacher_context} shows that working memory outperforms ground-truth-derived hints and their combination, whereas random-mask hints degrade performance.
We also compare forward KL, reverse KL, and JSD in Appendix~\ref{app:distillation_objective}; JSD achieves the best overall performance and is adopted as our default distillation objective.

\subsection{Qualitative Analysis}
\begin{figure*}[t]
\centering
\includegraphics[width=1\textwidth]{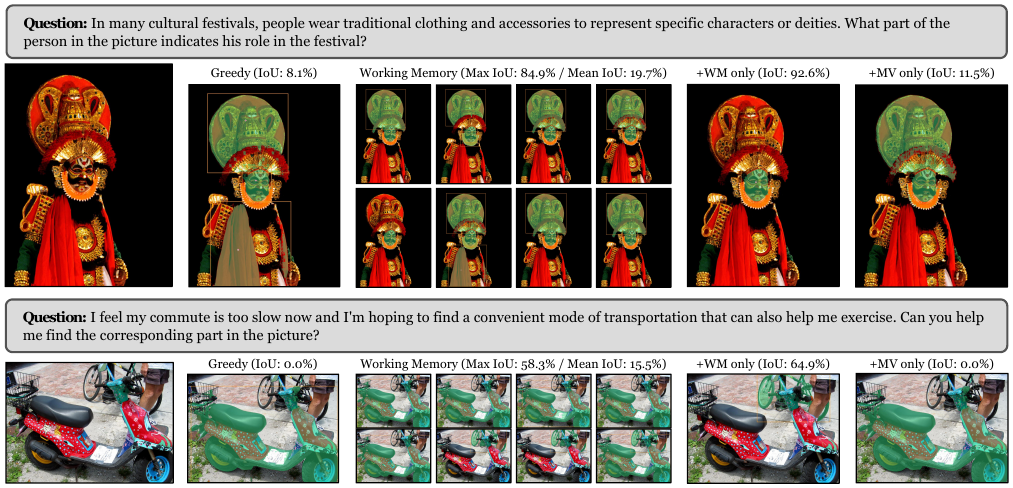}
\vspace{-3mm}
\caption{
\textbf{Working memory versus majority voting on ReasonSeg.}
Each row shows a query, the input image, a greedy prediction, eight sampled candidate masks, a WM-guided prediction, and an MV-only prediction.
WM-guided inference revisits the image and query using the corresponding responses as working memory, while MV directly aggregates the same eight masks.
}
\label{fig:wm_vs_majority_voting}
\label{fig:wm_qualitative}
\vspace{-1mm}
\end{figure*}

To illustrate the advantages of working-memory-guided inference~(WM) over majority voting (MV)~\citep{star}, we present two representative ReasonSeg examples in Figure~\ref{fig:wm_vs_majority_voting}.
All predictions are generated by the same \ours model trained from Qwen3-VL-8B.
We include greedy prediction as a reference, generating a single response from the image and query by selecting the highest-probability token at each decoding step.
For each example, WM-guided inference revisits eight sampled responses to generate a new prediction, while MV-only inference directly aggregates the masks predicted by those same responses without using working memory.

In the first example, the query asks which part of the person indicates their role in the festival.
Most sampled predictions include the large headdress, while the annotated target is the facial region.
MV preserves this error, whereas WM-guided inference localizes the face, achieving 92.6\% IoU compared with 11.5\% for MV.
In the second example, the query asks for a convenient means of transportation that also provides exercise.
Most sampled responses select the foreground scooter, overlooking the exercise requirement.
MV consequently retains the scooter, whereas revisiting these responses with working memory leads the model to identify the bicycle, which satisfies both requirements.

In both examples, WM-guided inference corrects target-selection errors that persist under majority voting and produces predictions with higher IoU than the best individual candidate.
These observations highlight the effectiveness of working memory in \ours, enabling the model to revisit prior attempts, refine its interpretation of the query, and localize the target more accurately.

\FloatBarrier

\section{Conclusion}

In this work, we presented \ours, a working-memory distillation framework for reasoning segmentation that enables a policy to learn from revisiting its own prior attempts.
By retaining reasoning traces and localization proposals from quality-selected rollouts, \ours constructs working memory that provides additional context for a self-teacher.
Joint optimization of on-policy self-distillation and outcome-based reinforcement learning transfers this contextual guidance into the student through complementary token-level supervision and segmentation feedback.
The resulting policy supports prediction from the image and query alone, while retaining the ability to revisit the input with working memory at inference time.
Experiments on reasoning segmentation benchmarks demonstrate the effectiveness of our approach, highlighting self-generated working memory as a useful source of supervision for improving reasoning and localization.

\clearpage
\bibliography{iclr2027_conference}
\bibliographystyle{iclr2027_conference}

\newpage
\appendix

\section{Implementation Details}
\label{app:implementation}

\paragraph{Architecture and optimization.}
We use Qwen3-VL-8B/32B-Instruct as the multimodal language model and SAM2.1 Hiera-Large as the mask generator. 
The vision backbone and SAM2.1 are frozen. 
We adapt the language model with LoRA~\citep{lora} on its linear layers, using rank 64 and scaling factor 64. 
The model predicts a list of boxes, points, and short labels. 
SAM2.1 converts the spatial prompts into object masks, which are combined to obtain the query-level prediction. We use a weight decay~\citep{adamw} of 0.001, zero entropy regularization, gradient checkpointing, and a random seed of 42 in the release configuration.

\paragraph{RL-only Warm-up (Stage~1).}
\textcolor{black}{Following StAR's Stage-1 data construction~\citep{star}, we use 5,166 training examples from LVIS, RefCOCOg, and gRefCOCO.}
\textcolor{black}{Neither \ours nor our reproduced StAR baseline uses the training split of ReasonSeg or ReasonSeg-X.}
Stage~1 runs for one epoch. 
The learning rate is $10^{-5}$, the query batch size is 24, and each query has 16 sampled responses. 
Training mixes ordinary queries and queries augmented with working memory. 
Working memory construction uses eight preliminary attempts. 
This stage teaches the model to reason about the image both with and without additional attempts in its context.

\paragraph{Working Memory
Distillation (Stage~2).}
In this stage, we optimize the joint objective $ \mathcal L_{\text{\ours}}=\mathcal L_{\text{GRPO}}+\lambda\mathcal L_{\text{OPSD}}$. 
Stage~2 uses 869 examples selected from the 5,166-example Stage~1 training pool, as described in Appendix~\ref{app:data_construction}.
The consolidated training recipe runs for two epochs with a constant learning rate of $10^{-6}$ and a query batch size of 16. 
For each query, $N=16$ attempts are sampled, and the top $K=8$ according to training mask IoU form the teacher's working memory. 
The student uses the ordinary context, while the teacher evaluates the student's generated prefixes with the self-generated working memory. 
Both the student and the teacher use the current model parameters. 
Rollout tensor parallelism is 2, and the actor micro-batch size is 2. 
The maximum training response length is 2,048 tokens.

\paragraph{Training hardware.}
We train both stages of \ours on 24 NVIDIA H800 GPUs.

\paragraph{Distributional supervision.}
The implementation computes JSD on a compressed vocabulary partition. 
It retains the teacher's top-16 tokens at each position and groups the remaining probability mass into one residual bin for both teacher and student. 
Probabilities on the retained tokens are normalized over the full vocabulary before this grouping. 
The loss therefore compares distributions on the same partition, rather than renormalizing only the retained tokens.

\paragraph{Inference and mask generation.}
The default evaluation uses greedy generation with a maximum of 1,024 new tokens. 
The LoRA~\citep{lora} adapter is loaded without merging. 
The image supplied to the language model is resized to $840\times840$, while the spatial prompts are mapped back to the original image geometry for SAM2.1. 
At test time, the ground-truth mask is not available to construct working memory. 
Instead, working memory contains unranked attempts sampled for the same image and query. 
The default working memory size~($K$) is 8. 
Multiple independently conditioned outputs can also be aggregated by mask voting. 
The main WM+MV configuration combines 32 WM-conditioned responses and 32 plain responses without working
memory.

\section{Prompts}
\label{app:prompts}
We provide the instruction text used by \ours. 
\texttt{Question} is replaced by the dataset query and \texttt{Answer} by the output-format example. 
The image is supplied through the model's multimodal input. 

\begin{tcolorbox}[title=RL-only warm-up~(stage~1) and referring expression segmentation,colback=gray!4,colframe=gray!50,breakable,fontupper=\small]
Please find "\{Question\}" with bbox(es) and point(s). Also provide a short label for each object. Compare the difference between object(s) and find the most closely matched object(s). Return ALL matching instances; double-check none are missed. Output the thinking process first, then the final answer in \texttt{<answer> </answer>} tags. Output the bbox(es) and point(s) inside the interested object(s), along with a short label, in JSON format. i.e., thinking process (step-by-step reasoning) here \texttt{<answer>\{Answer\}</answer>}
\end{tcolorbox}
\begin{tcolorbox}[title=Working memory distillation~(stage~2) and reasoning segmentation,colback=gray!4,colframe=gray!50,breakable,fontupper=\small]
Please find "\{Question\}" with bbox(es) and point(s). Also provide a short label for each object. First, understand and summarize what the query---"\{Question\}"---is likely referring to (which object or concept). Then apply this to the image and find the matched target object(s). Return ALL matching instances; if there are no matches, return an empty list (\texttt{<answer>[]</answer>}). double-check none are missed. Output the thinking process first, then the final answer in \texttt{<answer> </answer>} tags. Output the bbox(es) and point(s) inside the interested object(s), along with a short label, in JSON format. i.e., thinking process (step-by-step reasoning) here \texttt{<answer>\{Answer\}</answer>}
\end{tcolorbox}

\begin{tcolorbox}[title=Working-memory context appended to the query,colback=gray!4,colframe=gray!50,breakable,fontupper=\small]
Privileged hint:\par
Below are \{n\} previous attempts at answering this exact query. Use them to calibrate your answer --- identify patterns in what works and what doesn't. Answer exactly as you would without them: think step by step, then give \texttt{<answer>} with bbox(es) and point(s) in JSON format. Never mention these attempts in your response.\par\medskip
[Attempt 1] \{text\}\par
$\ldots$\par
[Attempt n] \{text\}
\end{tcolorbox}
\begin{tcolorbox}[title=JSON format example used by the plain evaluator,colback=gray!4,colframe=gray!50,breakable,fontupper=\small\ttfamily\raggedright]
[\{"label": "chair", "bbox\_2d": [10,100,200,210], "point\_2d": [30,110]\},
\{"label": "train track", "bbox\_2d": [225,296,706,786], "point\_2d": [302,410]\}]
\end{tcolorbox}

\section{Working Memory Distillation Training Set Construction}
\label{app:data_construction}
Working memory distillation (stage~2) focuses training on examples with different forms of remaining difficulty. 
We start from the 5,166-example RL-only warm-up (stage~1) training pool and generate eight plain responses per example using the stage~1 model. 
Each response is converted to a mask and scored against the training annotation. 
Let $u_1,\ldots,u_8$ denote these IoUs. 
We record their mean $\bar u$, maximum $u_{\max}$, minimum $u_{\min}$, the counts $n_{0.6}$ and $n_{0.8}$ at or above the corresponding thresholds, and the gap $\Delta=u_{\max}-\bar u$. These statistics distinguish inconsistent predictions from uniformly difficult or already reliable examples.
\begin{itemize}[leftmargin=*]
\item \textbf{Frontier examples} have $1\leq n_{0.6}\leq7$, no parsing errors, and $\Delta>0.1$. Some attempts succeed, but performance remains inconsistent. The final subset contains 364 such examples.
\item \textbf{Refinement examples} satisfy $n_{0.6}=8$ and $1\leq n_{0.8}\leq7$. They are broadly localized correctly but still leave room for mask-level improvement. We prioritize examples whose mean IoU is near 0.8 and retain 200.
\item \textbf{Hard but partially solvable examples} have $n_{0.6}=0$ and $u_{\max}\geq0.5$. We prioritize maximum IoUs near 0.6 and retain 97.
\item \textbf{Consistently unsuccessful examples} have $u_{\max}<0.1$. The final set includes 58 examples to retain exposure to difficult queries with no successful sampled attempt.
\item \textbf{Replay examples} satisfy $u_{\min}>0.8$. We randomly select 150 using seed 42 to retain examples on which the model is already reliable.
\end{itemize}
After validating the selected records, the final manifest contains 869 examples: 422 from LVIS, 265 from RefCOCOg~\citep{refcoco}, and 182 from gRefCOCO~\citep{grefcoco}. 
We preserve the original images, queries, and masks. 
All selection statistics are computed on training data. 

\section{Results on Additional Benchmarks}
\label{app:additional_benchmarks}

To further assess the generalization of \ours, we follow StAR~\citep{star} and evaluate on three additional settings: MUSE~\citep{pixellm} for multi-target segmentation, MMR~\citep{mmr} for compositional object-and-part segmentation, and the RefCOCO family~\citep{refcoco} for referring expression segmentation.
We report gIoU and cIoU on MUSE val and MMR val/test, and cIoU on RefCOCO testA, RefCOCO+ testA, and RefCOCOg test.

\begin{table*}[t]
\begin{minipage}[t]{0.37\textwidth}
\vspace{0pt}
\centering
\footnotesize
\setlength{\tabcolsep}{3pt}
\caption{\textbf{Performance on MUSE val.}
Gray rows use the original evaluation protocol and are shown for reference.}
\label{tab:muse_extra}
\resizebox{\linewidth}{!}{%
\begin{tabular}{llcc}
\toprule
\multirow{2}{*}{\textbf{Method}}
& \multirow{2}{*}{\shortstack{\textbf{Base model}\\\textbf{(Size)}}}
& \multicolumn{2}{c}{\textbf{MUSE val}} \\
\cmidrule(lr){3-4}
& & gIoU & cIoU \\
\midrule
\textcolor{gray}{LISA}
& \textcolor{gray}{LLaVA-7B}
& \textcolor{gray}{42.0} & \textcolor{gray}{46.1} \\
\textcolor{gray}{PixelLM}
& \textcolor{gray}{LLaVA-7B}
& \textcolor{gray}{42.6} & \textcolor{gray}{50.7} \\
\textcolor{gray}{POPEN}
& \textcolor{gray}{LLaVA-7B}
& \textcolor{gray}{45.4} & \textcolor{gray}{55.2} \\
\textcolor{gray}{LISA}
& \textcolor{gray}{LLaVA-13B}
& \textcolor{gray}{43.6} & \textcolor{gray}{50.2} \\
\textcolor{gray}{PixelLM}
& \textcolor{gray}{LLaVA-13B}
& \textcolor{gray}{44.8} & \textcolor{gray}{54.1} \\
\textcolor{gray}{POPEN}
& \textcolor{gray}{LLaVA-13B}
& \textcolor{gray}{48.0} & \textcolor{gray}{59.1} \\
\midrule
VisionReasoner & Qwen2.5-VL-7B & 50.5 & 47.6 \\
\midrule
StAR & Qwen3-VL-8B & 55.7 & 54.2 \\
+ MV & Qwen3-VL-8B & 57.1 & 55.5 \\
\midrule
\textbf{\ours} & Qwen3-VL-8B
& \underline{57.2} & \underline{55.7} \\
+ WM\&MV & Qwen3-VL-8B
& \textbf{58.3} & \textbf{56.8} \\
\bottomrule
\end{tabular}%
}

\end{minipage}\hfill
\begin{minipage}[t]{0.6\textwidth}
\vspace{0pt}
\centering
\footnotesize
\setlength{\tabcolsep}{3pt}
\caption{
\textbf{Performance on MMR val and test.}
Gray entries ($*$) use MMR training data and are shown for reference.
}
\label{tab:mmr_extra}
\resizebox{\linewidth}{!}{%
\begin{tabular}{llcccc}
\toprule
\multirow{2}{*}{\textbf{Method}}
& \multirow{2}{*}{\shortstack{\textbf{Base model}\\\textbf{(Size)}}}
& \multicolumn{2}{c}{\textbf{val}}
& \multicolumn{2}{c}{\textbf{test}} \\
\cmidrule(lr){3-4}\cmidrule(lr){5-6}
& & gIoU & cIoU & gIoU & cIoU \\
\midrule
LISA & Llama2-13B & 15.4 & 20.0 & 16.1 & 19.8 \\
\textcolor{gray}{LISA$^*$}
& \textcolor{gray}{Llama2-13B}
& \textcolor{gray}{22.3}
& \textcolor{gray}{33.4}
& \textcolor{gray}{23.0}
& \textcolor{gray}{29.2} \\
\textcolor{gray}{M$^2$SA$^*$}
& \textcolor{gray}{LLaVA-7B}
& \textcolor{gray}{27.8}
& \textcolor{gray}{48.6}
& \textcolor{gray}{30.9}
& \textcolor{gray}{46.8} \\
\textcolor{gray}{M$^2$SA$^*$}
& \textcolor{gray}{Llama2-13B}
& \textcolor{gray}{28.4}
& \textcolor{gray}{49.1}
& \textcolor{gray}{31.6}
& \textcolor{gray}{47.6} \\
VisionReasoner & Qwen2.5-VL-7B
& 26.7 & 21.4 & 28.4 & 21.7 \\
\midrule
StAR & Qwen3-VL-8B
& 29.4 & 26.6 & 31.9 & 27.4 \\
+ MV & Qwen3-VL-8B
& \underline{30.6} & 27.5 & \underline{32.9} & 28.1 \\
\midrule
\textbf{\ours} & Qwen3-VL-8B
& 30.3 & \underline{27.7} & 32.8 & \underline{28.5} \\
+ WM\&MV & Qwen3-VL-8B
& \textbf{31.1} & \textbf{28.1}
& \textbf{33.6} & \textbf{28.9} \\
\bottomrule
\end{tabular}%
}

\end{minipage}
\end{table*}

\begin{table}[t]
\centering
\setlength{\tabcolsep}{9pt}
\renewcommand{\arraystretch}{0.9}
\caption{
\textbf{Performance on referring expression segmentation benchmarks.}
We report cIoU results.
}
\label{tab:refcoco_extra}
\begin{tabular}{lccc}
\toprule
\multirow{2}{*}{\textbf{Method}}
& \textbf{RefCOCO} & \textbf{RefCOCO+} & \textbf{RefCOCOg} \\
& testA & testA & test \\
\midrule
LISA-7B & 76.5 & 67.4 & 68.5 \\
SegLLM & 81.5 & 73.0 & 73.6 \\
READ & 80.2 & 73.7 & 71.4 \\
CoReS & 78.6 & 70.0 & 70.7 \\
CoPRS-7B & 85.3 & 80.3 & 76.2 \\
SAM-R1 & 79.2 & 74.7 & 73.1 \\
SAM-Veteran & 80.8 & 76.6 & 73.4 \\
VisionReasoner & 76.6 & 72.1 & 67.2 \\
SAM 3 Agent-7B & 64.3 & 57.0 & 58.8 \\
SAM 3 Agent-72B & 74.9 & 70.8 & 70.2 \\
\midrule
StAR-8B & 77.4 & 73.0 & 73.1 \\
+ MV & 78.0 & 73.4 & 73.4 \\
\midrule
\rowcolor{gray!10}
\textbf{SWiM-8B} & 79.0 & 75.6 & 73.5 \\
\rowcolor{gray!10}
+ WM\&MV & 79.2 & 76.5 & 73.9 \\
\bottomrule
\end{tabular}
\end{table}

\paragraph{Multi-target and compositional object-and-part segmentation.}
As shown in Tables~\ref{tab:muse_extra} and~\ref{tab:mmr_extra}, \ours generally outperforms the reproduced StAR baseline on MUSE and both MMR splits, even without working memory at inference.
Combining working memory with majority voting further improves gIoU and cIoU across these evaluations.
These results demonstrate the generalization of \ours beyond reasoning segmentation to multi-target and compositional object-and-part segmentation.

\paragraph{Referring expression segmentation.}
Table~\ref{tab:refcoco_extra} presents the results on referring expression segmentation (RES) benchmarks.
Although our training uses only about 1.8k examples from the RefCOCOg training split, \ours outperforms the reproduced StAR baseline and remains competitive with existing methods, particularly VisionReasoner and SAM 3 Agent.
With working memory and majority voting, \ours achieves results comparable to SAM-R1, SAM-Veteran, and SegLLM, which make more extensive use of RefCOCOg training data.
These results suggest that \ours remains effective across RES tasks with varying reasoning demands.

\section{Detailed Results of Progressive Configurations}
\label{app:progression}
\begin{table*}[t]
\centering
\caption{
\textbf{Progressive evaluation of the core components of \ours.}
We report results on RS, RS-R, and RS-X using the Qwen3-VL-8B backbone.
}
\label{tab:swim_progression}
\begin{tabular}{@{}cccccccccccc@{}}
\toprule
\multicolumn{2}{c}{Training}
& \multicolumn{2}{c}{Inference}
& \multicolumn{2}{c}{RS}
& \multicolumn{2}{c}{RS-R}
& \multicolumn{2}{c}{RS-X}
& \multicolumn{2}{c}{Average} \\
\cmidrule(lr){1-2}\cmidrule(lr){3-4}
\cmidrule(lr){5-6}\cmidrule(lr){7-8}
\cmidrule(lr){9-10}\cmidrule(lr){11-12}
WM & OPSD & WM & MV
& gIoU & cIoU & gIoU & cIoU
& gIoU & cIoU & gIoU & cIoU \\
\midrule
& & & &
68.6 & 60.2 & 71.3 & 65.7
& 54.0 & 49.2 & 64.6 & 58.4 \\
$\checkmark$ & & & &
69.6 & 64.4 & 72.4 & 67.6
& 55.1 & 50.0 & 65.7 & 60.6 \\
$\checkmark$ & $\checkmark$ & & &
70.8 & 65.8 & \underline{73.0} & \textbf{70.8}
& 55.6 & 49.3 & 66.5 & 62.0 \\
$\checkmark$ & $\checkmark$ & $\checkmark$ & &
\underline{71.1} & \underline{67.0} & 72.7 & \textbf{70.8}
& \underline{57.2} & \underline{51.0}
& \underline{67.0} & \underline{62.9} \\
\rowcolor{gray!15}
$\checkmark$ & $\checkmark$ & $\checkmark$ & $\checkmark$ &
\textbf{71.6} & \textbf{67.3} & \textbf{73.5} & \underline{70.3}
& \textbf{58.1} & \textbf{51.5}
& \textbf{67.8} & \textbf{63.0} \\
\bottomrule
\end{tabular}
\end{table*}

Table~\ref{tab:swim_progression} provides the per-benchmark results underlying the progressive comparison in Figure~\ref{fig:swim_progression}.
The configurations isolate the incremental contributions of introducing working memory during RL warm-up, applying working-memory distillation, and incorporating working memory and majority voting at inference.
All configurations use the Qwen3-VL-8B backbone.
The full configuration achieves the highest average gIoU and cIoU, validating the effectiveness of integrating the core designs of \ours.

\section{Effect of Teacher Context}
\label{app:teacher_context}
\begin{table*}[t]
\centering
\caption{
\textbf{Comparison of teacher context configurations.}
WM denotes working memory, GT denotes ground-truth visual hints, and Random denotes surrogate annotations generated by SAM2 from random point prompts.
}
\label{tab:hint_configurations}
\begin{tabular}{ccccccccccc}
\toprule
\multirow{2}{*}{WM}
& \multirow{2}{*}{GT}
& \multirow{2}{*}{Random}
& \multicolumn{2}{c}{RS}
& \multicolumn{2}{c}{RS-R}
& \multicolumn{2}{c}{RS-X}
& \multicolumn{2}{c}{Average} \\
\cmidrule(lr){4-5}\cmidrule(lr){6-7}
\cmidrule(lr){8-9}\cmidrule(lr){10-11}
& & & gIoU & cIoU & gIoU & cIoU
& gIoU & cIoU & gIoU & cIoU \\
\midrule
& & $\checkmark$
& 55.4 & 52.4 & 56.0 & 56.0
& 43.0 & 37.7 & 51.5 & 48.7 \\
\rowcolor{gray!15}
$\checkmark$ & &
& \textbf{70.8} & \underline{65.8}
& \underline{73.0} & \textbf{70.8}
& \textbf{55.6} & \textbf{49.3}
& \textbf{66.5} & \textbf{62.0} \\
& $\checkmark$ &
& \underline{70.7} & 65.5
& \textbf{73.2} & 70.1
& \underline{54.4} & \underline{48.3}
& 66.1 & 61.3 \\
$\checkmark$ & $\checkmark$ &
& \underline{70.7} & \textbf{67.1}
& 72.6 & \underline{70.2}
& \textbf{55.6} & 48.2
& \underline{66.3} & \underline{61.8} \\
\bottomrule
\end{tabular}
\end{table*}

To examine how teacher context affects distillation, we compare working memory (WM), ground-truth visual hints (GT), their combination, and random visual hints.
The random hints use SAM2 masks generated from random point prompts as surrogate annotations.

As shown in Table~\ref{tab:hint_configurations}, working memory alone achieves the highest average gIoU and cIoU.
GT hints improve some individual metrics but provide no overall advantage over working memory, either alone or in combination with it.
These results show that providing the teacher with more direct target information does not necessarily produce more effective supervision for the student.
In our setting, the student learns from the teacher's next-token distributions while receiving only the original image and query.
The value of teacher context therefore lies in the guidance transferred to the student, rather than the explicitness of the target information it contains.
Working memory provides effective guidance by allowing the teacher to revisit the policy's own reasoning traces and localization proposals, achieving better average student performance than annotation-based context.

Random hints yield substantially lower scores, further emphasizing the importance of relevant teacher context.
Overall, these results validate working memory as an effective source of supervision for distillation.

\section{Distillation Objective for OPSD}
\label{app:distillation_objective}
\begin{table}[t]
\centering
\caption{
\textbf{Comparison of OPSD divergence objectives.}
All variants use working memory as the teacher context during training and are evaluated without working memory or majority voting.
}
\label{tab:opd_loss_ablation}
\begingroup
\small
\setlength{\tabcolsep}{5pt}
\renewcommand{\arraystretch}{1.12}
\begin{tabular}{lcccccccc}
\toprule
\multirow{2}{*}{Loss}
& \multicolumn{2}{c}{RS}
& \multicolumn{2}{c}{RS-R}
& \multicolumn{2}{c}{RS-X test}
& \multicolumn{2}{c}{Average} \\
\cmidrule(lr){2-3}\cmidrule(lr){4-5}
\cmidrule(lr){6-7}\cmidrule(lr){8-9}
& gIoU & cIoU & gIoU & cIoU
& gIoU & cIoU & gIoU & cIoU \\
\midrule
Reverse KL
& \underline{70.5} & \underline{64.4}
& \textbf{73.1} & \textbf{70.8}
& \underline{54.0} & 46.0
& \underline{65.9} & 60.4 \\
Forward KL
& 70.2 & 63.9
& \textbf{73.1} & \underline{70.4}
& 53.8 & \underline{48.6}
& 65.7 & \underline{61.0} \\
\rowcolor{gray!10}
JSD
& \textbf{70.8} & \textbf{65.8}
& \underline{73.0} & \textbf{70.8}
& \textbf{55.6} & \textbf{49.3}
& \textbf{66.5} & \textbf{62.0} \\
\bottomrule
\end{tabular}
\endgroup
\end{table}

To examine the effect of the divergence used in OPSD, we compare forward KL, reverse KL, and JSD.
All variants use working memory as the additional teacher context and are evaluated using only the original image and query, without working memory or majority voting.

As shown in Table~\ref{tab:opd_loss_ablation}, JSD achieves the highest average gIoU and cIoU, with the best results on both RS and RS-X.
Forward and reverse KL attain slightly higher gIoU on RS-R, while reverse KL matches JSD in cIoU on this benchmark.
Overall, JSD provides the strongest aggregate performance across the three benchmarks, supporting its use as the default distillation objective in \ours.

\section{Sensitivity of the Distillation Weight}
\label{app:distillation_weight}
Table~\ref{tab:jsd_weight} provides the per-benchmark results for the distillation-weight analysis in Figure~\ref{fig:parameter_sensitivity}.
We vary $\lambda$ to examine the balance between OPSD and GRPO, with all configurations evaluated without working memory at inference.

Among the evaluated settings, $\lambda=0.1$ achieves the highest average gIoU and cIoU.
Larger weights improve some individual metrics but do not yield better overall performance, and increasing $\lambda$ to $10$ lowers both average scores relative to the default.
These results support a moderate contribution from distillation alongside outcome-based reinforcement learning and motivate our choice of $\lambda=0.1$.

\begin{table}[t]
\centering
\caption{
\textbf{Effect of the distillation weight $\lambda$.}
All settings are evaluated without working memory.
}
\label{tab:jsd_weight}
\small
\setlength{\tabcolsep}{4.5pt}
\begin{tabular}{@{}ccccccccc@{}}
\toprule
\multirow{2}{*}{$\lambda$}
& \multicolumn{2}{c}{RS}
& \multicolumn{2}{c}{RS-R}
& \multicolumn{2}{c}{RS-X}
& \multicolumn{2}{c}{Average} \\
\cmidrule(lr){2-3}\cmidrule(lr){4-5}
\cmidrule(lr){6-7}\cmidrule(lr){8-9}
& gIoU & cIoU & gIoU & cIoU
& gIoU & cIoU & gIoU & cIoU \\
\midrule
\rowcolor{gray!10}
0.1
& \textbf{70.8} & \textbf{65.8}
& \textbf{73.0} & \underline{70.8}
& \textbf{55.6} & 49.3
& \textbf{66.5} & \textbf{62.0} \\
0.5
& \underline{70.5} & 65.0
& 72.7 & 68.8
& \underline{55.0} & 48.5
& 66.1 & 60.8 \\
1.0
& \textbf{70.8} & 63.4
& \textbf{73.0} & \textbf{71.1}
& 54.4 & \underline{49.5}
& 66.1 & 61.3 \\
2.0
& \textbf{70.8} & \underline{65.7}
& \underline{72.9} & 69.9
& 54.9 & \textbf{49.9}
& \underline{66.2} & \underline{61.9} \\
10.0
& 70.3 & 65.1
& 72.4 & 69.6
& 54.0 & 48.0
& 65.6 & 60.9 \\
\bottomrule
\end{tabular}
\end{table}

\section{Working-Memory Size}
\begin{figure*}[t]
\centering\begin{minipage}[t]{0.40\textwidth}
\vspace{0pt}\centering
\includegraphics[width=0.97\linewidth]{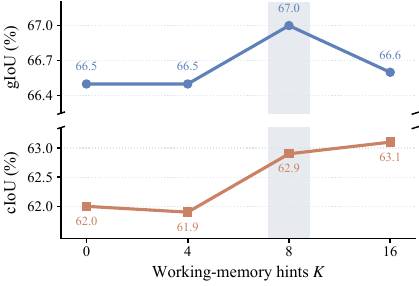}
\vspace{-2.5mm}
\caption{
\textbf{Effect of working-memory size at inference.}
Scores are averaged over RS, RS-R, and RS-X.
}
\label{fig:wm_size_sensitivity}
\end{minipage}\hfill\begin{minipage}[t]{0.56\textwidth}
\vspace{0pt}
\centering\def\WMHintCountInline{}
\ifdefined\WMHintCountInline
\makeatletter
\def\@captype{table}
\makeatother
\else
\begin{table*}[t]
\fi
\centering
\caption{
\textbf{Effect of working-memory size at inference.}
$K$ denotes the number of prior responses.
}
\label{tab:wm_hint_count}
\begingroup
\small
\setlength{\tabcolsep}{3.5pt}
\renewcommand{\arraystretch}{1.12}
\ifdefined\WMHintCountInline
\def\WMTableFit#1{\resizebox{\linewidth}{!}{#1}}
\else
\def\WMTableFit#1{#1}
\fi
\WMTableFit{%
\begin{tabular}{@{}lcccccccc@{}}
\toprule
\multirow{2}{*}{$K$}
& \multicolumn{2}{c}{RS}
& \multicolumn{2}{c}{RS-R}
& \multicolumn{2}{c}{RS-X}
& \multicolumn{2}{c}{Average} \\
\cmidrule(lr){2-3}\cmidrule(lr){4-5}
\cmidrule(lr){6-7}\cmidrule(lr){8-9}
& gIoU & cIoU & gIoU & cIoU
& gIoU & cIoU & gIoU & cIoU \\
\midrule
0
& 70.8 & 65.8
& \textbf{73.0} & \textbf{70.8}
& 55.6 & 49.3
& 66.5 & 62.0 \\
4
& \underline{70.9} & 66.7
& 72.2 & 68.6
& 56.5 & 50.5
& 66.5 & 61.9 \\
\rowcolor{gray!10}
8
& \textbf{71.1} & \underline{67.0}
& \underline{72.7} & \textbf{70.8}
& \textbf{57.2} & \underline{51.0}
& \textbf{67.0} & \underline{62.9} \\
16
& 70.8 & \textbf{69.0}
& 72.4 & \underline{69.2}
& \underline{56.6} & \textbf{51.1}
& \underline{66.6} & \textbf{63.1} \\
\bottomrule
\end{tabular}%
}
\endgroup
\ifdefined\WMHintCountInline
\else
\end{table*}
\fi

\end{minipage}
\end{figure*}

We examine the effect of working-memory size at inference time by varying the number of prior responses provided as context, with $K=0$ denoting plain inference.
Figure~\ref{fig:wm_size_sensitivity} summarizes the average performance, and Table~\ref{tab:wm_hint_count} provides the per-benchmark results.

Using eight prior responses improves both average metrics over plain inference and achieves the highest average gIoU.
Increasing the memory size to $K=16$ yields a further gain of 0.2 percentage points in average cIoU but reduces average gIoU.
Thus, expanding working memory does not consistently improve performance across metrics.
These results support $K=8$ as a practical choice, providing strong overall performance with fewer contextual responses than $K=16$.

\label{app:memory_size}

\section{Additional Qualitative Analysis}
\label{app:qualitative}
\begin{figure*}[p]
\centering
\includegraphics[width=\textwidth]{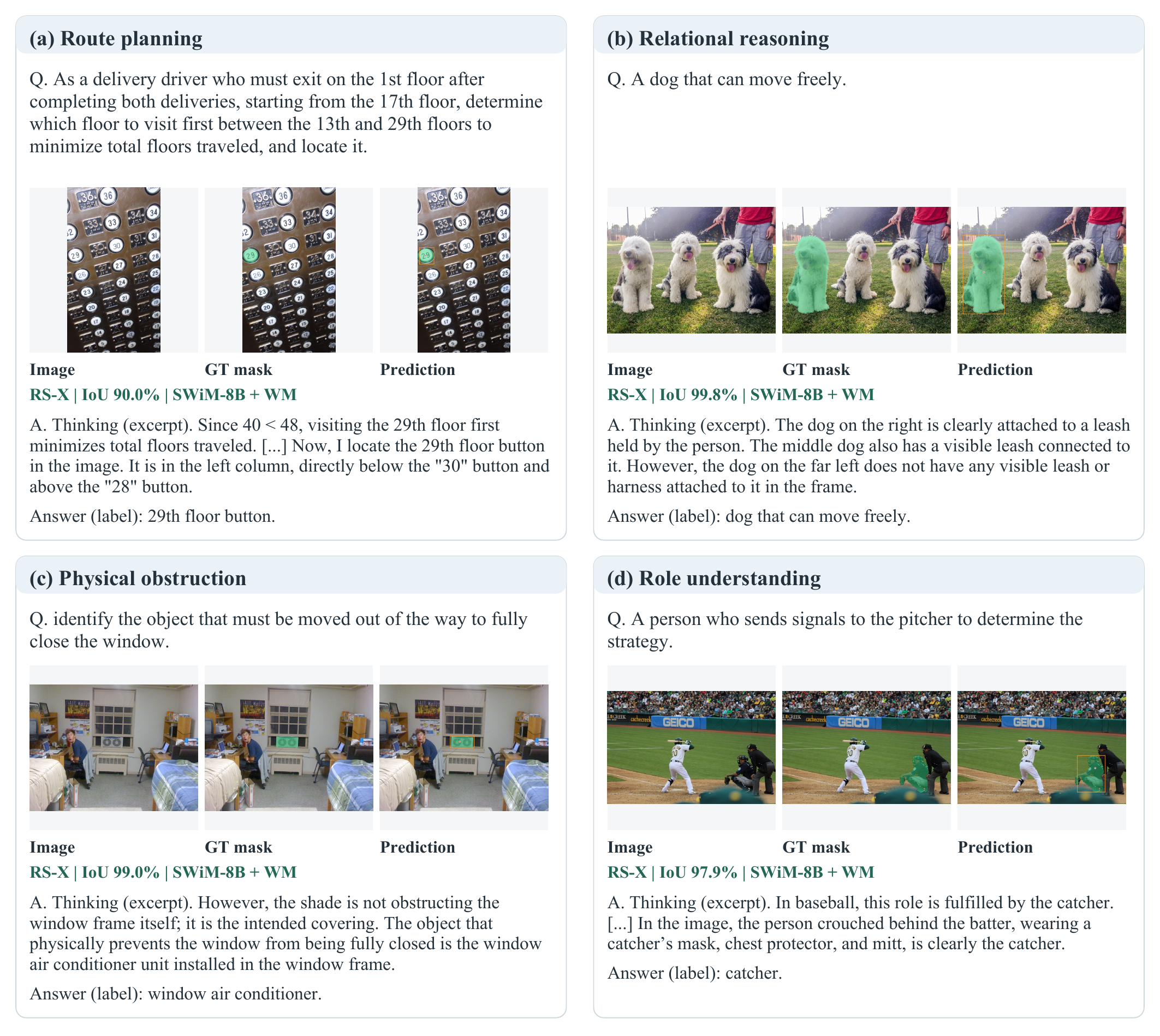}
\caption{\textbf{Question--answer visualizations on ReasonSeg-X.} 
Each panel shows the query, original image, ground-truth mask, and predicted mask with the model-generated box and point, followed by verbatim thinking excerpts and answer labels. 
Green overlays denote masks, orange rectangles denote predicted boxes, and pink markers denote predicted points. 
IoUs are per-example scores. 
The examples cover route planning, visible constraints on movement, physical obstruction, and role understanding.
}
\label{fig:additional_qualitative}
\label{fig:rsx_qa_success}
\end{figure*}

Figure~\ref{fig:additional_qualitative} presents four qualitative examples from ReasonSeg-X using \ours with the Qwen3-VL-8B backbone.
They illustrate route planning, reasoning about physical constraints, and recognition of functional roles.

In example (a), the model compares two delivery orders and chooses the 29th floor first, reducing the total travel from 48 to 40 floors before exiting on the first floor. 
In example (b), it compares the visible restraints on three dogs and selects the dog without a visible leash.
Example (c) distinguishes a window covering from an object that physically obstructs closing the window, identifying the installed air conditioner. 
In example (d), the model connects the role of communicating with the pitcher to the catcher and grounds this interpretation using the protective equipment and crouched posture.

\begin{figure*}[t]
\centering
\includegraphics[width=\textwidth]{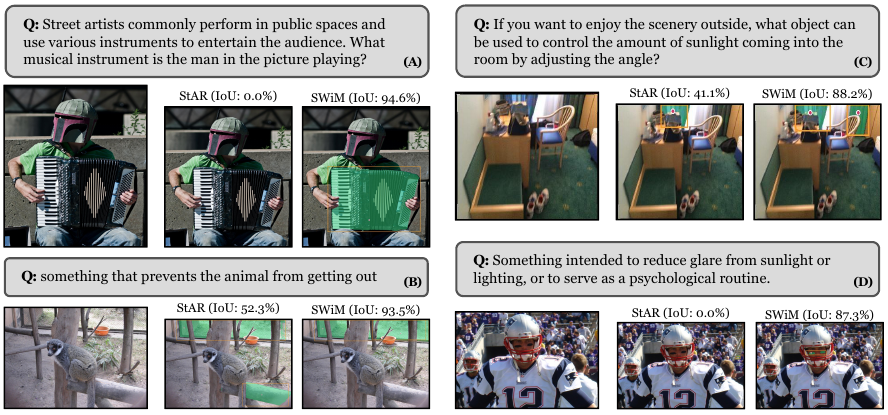}
\caption{
\textbf{Qualitative comparison with StAR on the ReasonSeg series.}
Each example shows the image and query alongside predictions from our reproduced StAR baseline and \ours, both using Qwen3-VL-8B.
These examples highlight improvements in target identification, region coverage, and small-target localization.
}
\label{fig:star_vs_swim_qualitative}
\end{figure*}

\paragraph{Comparison with StAR.}
Figure~\ref{fig:star_vs_swim_qualitative} presents a qualitative comparison between \ours and our reproduced StAR baseline on the ReasonSeg series, both using the Qwen3-VL-8B backbone.
Both models use plain greedy decoding with the same query template and input resolution, without working memory or majority voting at inference. 
In example (A), \ours identifies the accordion from the functional description, improving mask IoU from 0.0\% to 94.6\%. 
Example (B) shows more complete coverage of the queried enclosure, with IoU increasing from 52.3\% to 93.5\%. 
In example (C), \ours covers more of the annotated window-covering region. 
Example (D) further illustrates precise localization of a small target: \ours selects the eye-black strips described by the query. 
Together, these cases illustrate improvements in target selection and spatial coverage under plain inference, consistent with the goal of transferring memory-conditioned guidance into the model parameters.

\section{Failure Case Analysis}
\label{app:failures}
\begin{figure*}[p]
\centering
\includegraphics[width=\textwidth]{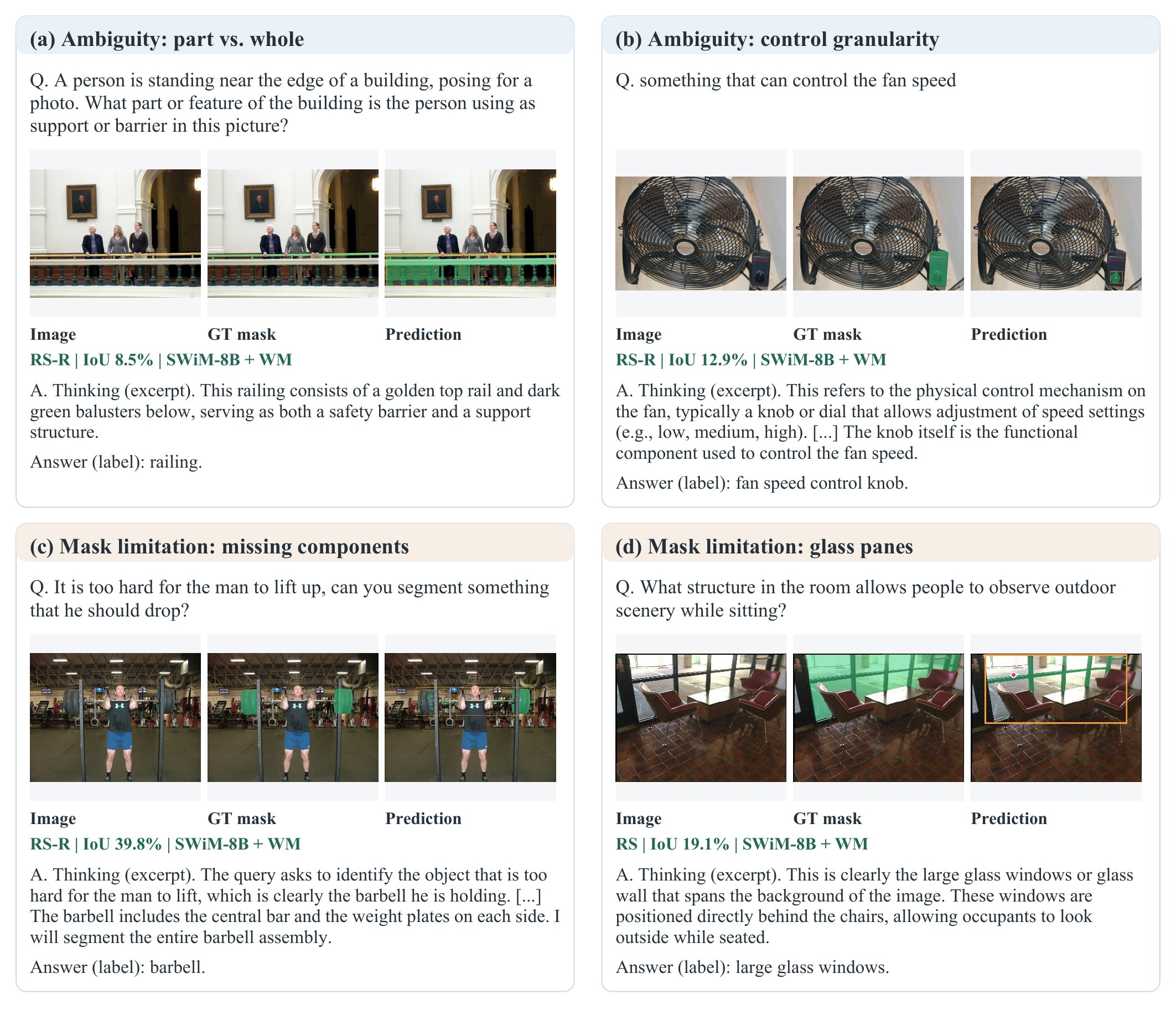}
\caption{
\textbf{Two types of low-IoU predictions on the ReasonSeg series.} 
All predictions are generated by \ours with the Qwen3-VL-8B backbone using working memory at inference.
The top row shows target-granularity ambiguities: handrail versus railing, and control unit versus knob. The bottom row shows incomplete SAM2 masks despite plausible object descriptions and localization prompts: a barbell and glass windows. Each panel includes the original image, GT mask, prediction with box and point, and verbatim thinking excerpts with answer labels. These examples diagnose possible error sources; they do not establish annotation errors or a causal attribution to the mask decoder.
}
\label{fig:failure_cases}
\label{fig:failure_qa_taxonomy}
\end{figure*}

Figure~\ref{fig:failure_cases} examines two sources of disagreement with the reference masks: ambiguity in the target extent and incomplete mask generation despite plausible localization. 

\paragraph{Ambiguity in target extent.} 
\textcolor{black}{
In examples (a) and (b), the queries may admit different plausible target extents, such as the handrail versus the railing and the fan control unit versus its knob; a single reference mask may therefore penalize a reasonable alternative. 
This motivates exploring multiple candidate masks for ambiguous queries and evaluating them against multiple human-validated reference masks.
}

\paragraph{Incomplete masks from plausible localization.}
\textcolor{black}{In examples (c) and (d), the MLLM policy identifies and localizes the intended targets, but SAM2 produces incomplete masks (39.8\% and 19.1\% IoU, respectively). Future work could let the policy provide more spatial prompts or jointly optimize the policy and SAM2.}

\FloatBarrier

\end{document}